\documentclass{article}

\PassOptionsToPackage{numbers,sort&compress}{natbib}
\usepackage{natbib}
\usepackage{arydshln}
\usepackage{algorithm}
\usepackage{algorithmic}
\usepackage{booktabs}
\usepackage{siunitx}
\usepackage[main, final]{neurips_2026}
\usepackage[utf8]{inputenc} % allow utf-8 input
\usepackage[T1]{fontenc}    % use 8-bit T1 fonts
\usepackage{hyperref}       % hyperlinks
\usepackage{url}            % simple URL typesetting
\usepackage{booktabs}       % professional-quality tables
\usepackage{amsfonts}       % blackboard math symbols
\usepackage{nicefrac}       % compact symbols for 1/2, etc.
\usepackage{microtype}      % microtypography
\usepackage{xcolor}         % colors

\usepackage{amsmath, amssymb}
\usepackage{graphicx}
\usepackage{multirow}
\usepackage{caption}
\usepackage{subcaption}
\usepackage[dvipsnames,table,xcdraw]{xcolor}
\title{DiA: Directional Adapter}

\author{%
  Ashish Kumar Singh$^{*}$ \\
  Machine Learning Group\\
  Luleå University of Technology, Sweden\\
  \texttt{ashish.kumar.singh@associated.ltu.se} \\
  \And
  Prakash Chandra Chhipa$^{*}$ \\
  Scalable Systems\\
  RISE Research Institutes of Sweden, Sweden\\
  \texttt{prakash.chandra.chhipa@ri.se} \\
  \AND
  \vspace{-0.8cm}
  \textit{$^{*}$Joint first authors with equal contributions.}
}

\begin{document}

\maketitle
\begin{abstract}
Fine-grained action recognition often depends on how an action unfolds over time rather than on appearance alone. Visually similar classes may differ in preparation, execution, or completion phases, making uniformly aggregated temporal context suboptimal. We study this problem under a practical objective: i) improving recognition performance with few tunable parameters, ii) being compute efficient, and iii) using only the readily available video modality. Motivated by this missing temporal observation and the accuracy--compute efficiency--unimodality objective, we propose \textbf{Directional Adapter (DiA)}, a parameter-efficient adapter on top of CLIP. DiA formulates temporal specialization into causal and anti-causal directions, allowing past-to-present and future-to-present cues to be modeled differently. To keep this directional modeling parameter- and compute-efficient, both directions use depth-wise temporal convolutions in a compact bottleneck space and are combined through a learnable fusion. DiA achieves state-of-the-art performance among methods using a single video input modality across seven public benchmarks, with only $\sim$3.6M tunable parameters, while also delivering higher throughput and lower inference latency than prior methods. Code is available at \url{https://github.com/prakashchhipa/DiA-Directional-Adapter}.
\end{abstract}

\section{Introduction}
\begin{figure}[t]
  \centering
  \includegraphics[width=1.0\linewidth]{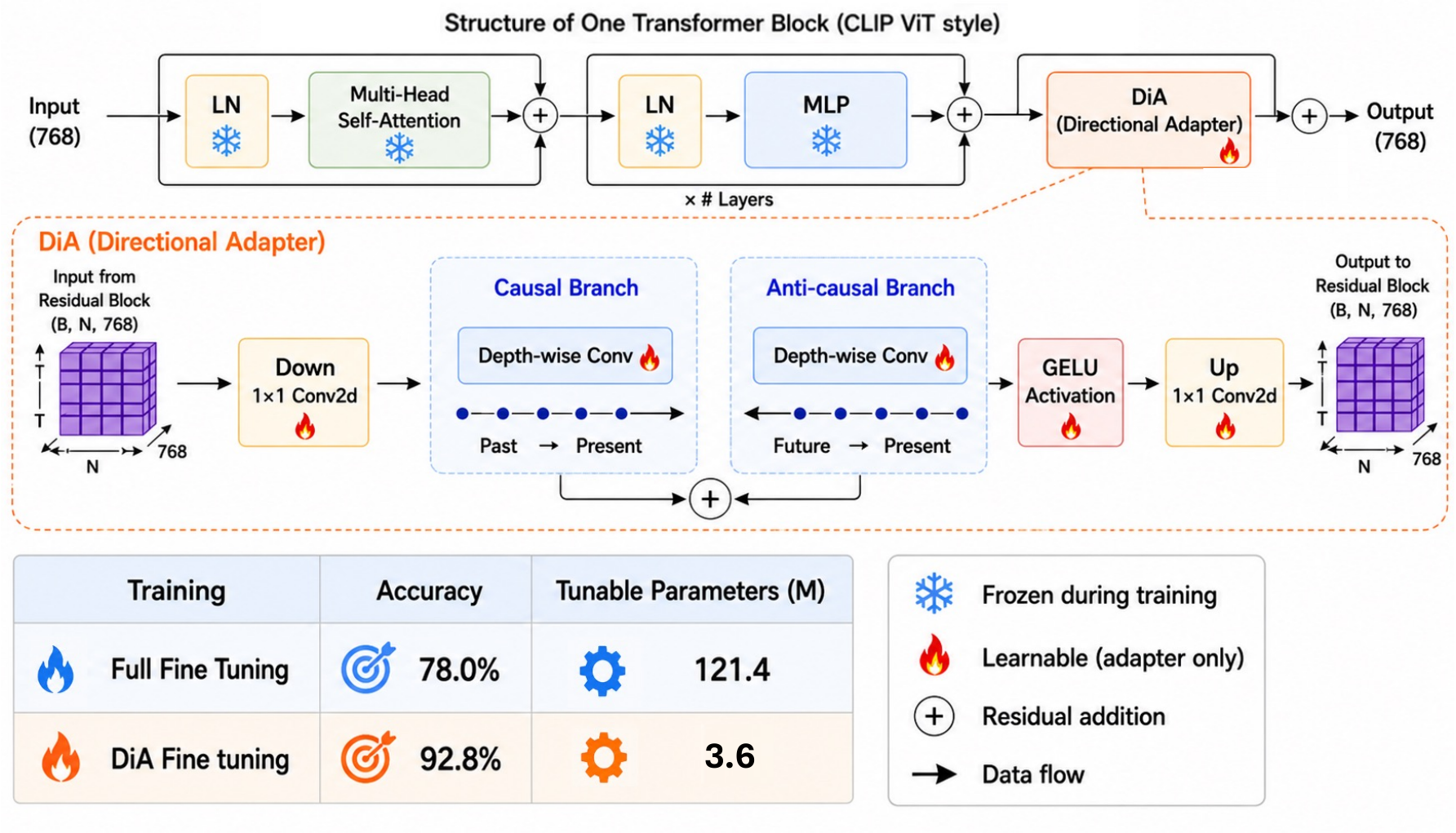}
  \caption{
  \textbf{Overview of DiA.}
DiA targets fine-grained action recognition under the accuracy--compute efficiency--unimodality objective. Compared with Full fine-tuning~\cite{bertasius2021space} and prior efficient adapters \cite{yang2023aim}, DiA introduces directional temporal specialization to a frozen CLIP visual backbone using lightweight causal and anti-causal adapters with learnable directional fusion.}
  \label{fig:overview_and_results}
\vspace{-6mm}
\end{figure}

Fine-grained action recognition requires more than detecting motion. Many action classes are visually similar in appearance and scene context, yet differ in the direction in which motion evidence accumulates over time. In benchmarks such as Diving48 \cite{li2018diving48} and FineGym \cite{shao2020finegym}, class boundaries are often determined by rotation count, transition direction, landing mechanics, body configuration, or short-lived execution stages rather than by objects or background context. The recognition problem is therefore temporal \cite{zhang2021temporal} and asymmetric: the \textit{temporal evidence} (discriminative features used in recognition) before an action state and the evidence after it are not interchangeable, and treating the full temporal context uniformly can weaken the cues that separate fine-grained classes.

Further, existing action recognition methods expose two practical challenges. First, full or large-scale fine-tuning of pretrained video models improves adaptation capacity but increases the number of tunable parameters and training cost, as in CLIP-based video methods such as X-CLIP~\cite{ni2022expanding} and TimeSformer ~\cite{bertasius2021space}. 
%Heavy video-specific architectures introduce a related compute burden: I3D~\cite{ao2017quo} and SlowFast~\cite{feichtenhofer2019slowfast} model spatiotemporal structure with 3D or multi-pathway computation. 
Thus, temporal modeling often comes at the cost of either more tunable parameters or higher compute. Second, multi-modal methods can improve fine-action representations, but they add modality-specific data and pre-processing requirements beyond the video modality. Video-text methods such as BDC-CLIP~\cite{long2025bdc} depend on language supervision or prompt construction, while existing fine-grained action recognition methods such as PeVL~\cite{zhang2024pevl} require pose estimation before recognition. 
For fine-grained closed-set action recognition, language cues may be weak because class names can be semantically close, whereas the decisive distinction lies in frame-level motion. For example, FineGym includes categories such as \textit{switch leap with 0.5 turn} versus \textit{switch leap with 1 turn}, while Diving48 includes categories such as \textit{Forward 25som 2Twis PIKE} versus \textit{Forward 25som 3Twis PIKE}. Additionally, pose estimation can also introduce an additional pre-processing dependency. Thus, multi-modality can add useful information, but it does not remove the need to recover discriminative temporal evidence from the video modality itself.

This leads to our research question:
\emph{What temporal formulation is needed to improve fine-grained action recognition using only the video modality, while remaining compute- and parameter-efficient?}

Therefore, we study fine-grained action recognition under the \textbf{accuracy--compute efficiency--unimodality objective}: i) improving recognition performance with few tunable parameters, ii) being compute efficient, and iii) using only the readily available video modality. This objective directly follows from the two challenges above. The model must recover asymmetric temporal evidence from video itself, while avoiding the cost of full fine-tuning, heavy video-specific computation, or additional modality pipelines.

We propose \textbf{DiA} (\textbf{Di}rectional \textbf{A}dapter) to address this question (refer Fig. \ref{fig:overview_and_results}). DiA is a parameter-efficient adapter inserted into a CLIP visual backbone. Its design follows directly from the problem structure. Fine-grained actions exhibit temporal asymmetry; therefore, DiA does not treat temporal context uniformly. Instead, it decomposes temporal evidence into causal (past-to-present) and anti-causal (future-to-present) directions. Since the relative importance of these directions can vary across actions, DiA exploits directional fusion \cite{zhu2022fencenet} with learnable capability. To satisfy the efficiency requirement, temporal specialization is implemented using depth-wise temporal convolutions in a compact bottleneck space and applied as a token-local residual update. This formulation preserves pretrained visual priors while introducing explicit directional temporal specialization for fine-grained action recognition.

Empirical results across seven public benchmarks demonstrate that DiA satisfies the accuracy--compute efficiency--unimodality objective, achieving state-of-the-art performance among single-stream methods using only the video modality. For instance, on Diving48, DiA achieves 92.8\% top-1 accuracy with only $\sim$3.0M tunable parameters, improving over prior CLIP-based adaptation while reducing computational cost by 31.24\% and increasing throughput by 2.17$\times$ compared to AIM~\cite{yang2023aim}. These results show that directional temporal specialization improves recognition performance while remaining parameter- and compute-efficient using a single video input stream.

Our contributions are:
\vspace{-4mm}
\begin{itemize}
    \item We formulate fine-grained action recognition as a \textbf{temporal specialization} problem on top of CLIP, targeting the accuracy--compute efficiency--unimodality objective.
    \item We adapt \textbf{directional temporal adaptation}, which decomposes temporal evidence into causal and anti-causal directions to model asymmetric past-to-present and future-to-present cues for \textbf{fine-grained action recognition}.
    \item We propose \textbf{learnable directional fusion}, enabling adaptive selection of complementary directional evidence across fine action dynamics.
\end{itemize}

\section{Related Work}
\label{sec:relwork}
%We discuss work most relevant to fine-grained action recognition.
\paragraph{Conventional video action recognition methods.}
Prior RGB action-recognition methods establish the importance of temporal modeling. TSN~\cite{wang2018temporal}, TSM~\cite{lin2019tsm}, TRN~\cite{zhou2018temporal}, and TQN~\cite{zhang2021temporal} model temporal evidence using segment aggregation, temporal shifts, relational reasoning, or temporal queries, while I3D~\cite{ao2017quo} and SlowFast~\cite{feichtenhofer2019slowfast} use 3D convolutions or multi-pathway designs. These approaches show that temporal structure matters, but they are not designed around the accuracy--compute efficiency--unimodality objective. They either rely on substantial video-specific computation or do not explicitly target the phase-sensitive directional cues that define fine-grained action boundaries.

\paragraph{CLIP-based video recognition methods.}
Visual foundation models provide a stronger starting point for this objective. CLIP~\cite{radford2021learning} offers transferable visual priors learned from large-scale pretraining, and CLIP-based video methods such as ActionCLIP~\cite{wang2021actionclip}, X-CLIP~\cite{ni2022expanding}, EVL~\cite{lin2022frozen}, and ViFi-CLIP~\cite{rasheed2023fine} adapt these priors to video recognition. However, full or large-scale adaptation increases tunable parameters and computational cost. Parameter-efficient methods such as ST-Adapter~\cite{pan2022st} and AIM~\cite{yang2023aim} improve this trade-off by preserving the pretrained backbone and adding lightweight temporal adaptation. Yet their temporal specialization are not explicitly formulated around the core fine-grained observation that preparation, execution, and completion phases may contribute different evidence, and that past-to-present and future-to-present context need not be symmetric.

A separate line of work enriches video representations with language, pose, or both, including BDC-CLIP~\cite{long2025bdc}, CLAVER~\cite{jingyikronecker}, and PeVL~\cite{zhang2024pevl}. These cues can be useful when auxiliary semantics or body structure are reliable. In fine-grained closed-set recognition, however, the main ambiguity often remains temporal: class names can be nearly identical, while the decisive evidence lies in motion order and phase evolution. Pose can also be unreliable under fast motion or occlusion; FineGym reports only 44.7\% detectable joints with AlphaPose~\cite{shao2020finegym}, and PoseConvC3D reaches 54.5\% top-1 accuracy on Diving48~\cite{duan2022revisiting}. Thus, relying on auxiliary streams can move the method away from the unimodality part of the objective without directly addressing the missing temporal observation.
\section{Directional Adapter}
\label{sec:dia}

% We propose \textbf{DiA} (\textbf{Di}rectional \textbf{A}dapter), a parameter-efficient temporal adapter for fully supervised fine-grained action recognition on top of a CLIP visual backbone. Fine-grained action categories often share highly similar appearance and scene context, while differing primarily in subtle temporal ordering, directional motion evolution, and short-lived phase transitions. Existing parameter-efficient video adaptation methods are largely designed for general action recognition and typically rely on symmetric temporal modeling or joint spatio-temporal adaptation, which does not explicitly address the directional and phase-sensitive temporal structure that separates fine action classes. DiA addresses this gap by performing compact temporal specialization in the bottleneck space, explicitly decomposing temporal information into \emph{causal} and \emph{anti-causal} directions and fusing them through a learnable directional weighting.

We propose \textbf{Di}rectional \textbf{A}dapter (\textbf{DiA}, refer Fig. \ref{fig:overview_and_results}), a parameter-efficient temporal adapter for fully supervised fine-grained action recognition using a CLIP visual backbone. Fine-grained action categories often share similar appearance and scene context, but differ in temporal order, motion evolution, and short-lived action phases. Existing parameter-efficient video adapters mainly model temporal context symmetrically or combine spatial and temporal adaptation, which can obscure these asymmetric cues. DiA addresses this by applying compact temporal specialization in a bottleneck space, with separate \emph{causal} and \emph{anti-causal} directions whose outputs are combined by learnable directional weights.

Let $B$ denote the batch size, $T$ the number of input frames, $H\times W$ the spatial grid of patch tokens per frame, and $C$ the channel dimension. Given the token tensor entering the adapter,
\begin{equation}
\mathbf{X} \in \mathbb{R}^{(HW+1)\times(BT)\times C},
\end{equation}
we separate the class token $\mathbf{x}_{\mathrm{cls}} \in \mathbb{R}^{1\times(BT)\times C}$ and the patch tokens $\mathbf{X}_{p}\in\mathbb{R}^{HW\times(BT)\times C}$ as
\begin{equation}
\mathbf{X} = [\mathbf{x}_{\mathrm{cls}}; \mathbf{X}_{p}].
\end{equation}
We then reshape the patch tokens into a spatio-temporal tensor
\begin{equation}
\mathbf{V}\in\mathbb{R}^{B\times C\times T\times H\times W}.
\label{eq:reshape_v}
\end{equation}

DiA is derived from four modeling hypotheses for fine-grained action recognition, where the discriminative evidence is often temporally directional and phase-sensitive:

\noindent\textbf{H1: Low-DoF temporal specialization.}
Effective supervised adaptation should remain capacity-controlled. We therefore adapt features in a bottleneck space of width
\begin{equation}
C_a = \lfloor \alpha C \rfloor,
\end{equation}
where $C_a$ is the adapter width and $\alpha\in(0,1]$ is the bottleneck ratio, and perform temporal modeling using depth-wise convolutions, which restrict the update to local temporal structure with low parameter cost.

\noindent\textbf{H2: Directional temporal asymmetry.}
Fine-grained actions are not temporally symmetric: evidence from the past and evidence from the future need not contribute equally to recognition. We therefore formulate temporal specialization into a \emph{causal} direction and an \emph{anti-causal} direction rather than using a single symmetric temporal specialization.

\noindent\textbf{H3: Adaptive directional evidence selection.}
The importance of past-oriented versus future-oriented context varies across datasets, layers, and action phases. Thus, their contributions should not be fixed manually; instead, we apply a learnable directional fusion.

\noindent\textbf{H4: Residual identity-preserving adaptation.}
A pretrained CLIP encoder already provides strong visual priors. To preserve these priors and stabilize optimization, the adapter should act as a residual update that is initialized near identity at the start of training.

\paragraph{Bottleneck temporal adaptation.}
We first compress the input features using a $1\times1$ projection:
\begin{equation}
\mathbf{U} = \phi_{\downarrow}(\mathbf{V}),
\qquad
\phi_{\downarrow}: \mathbb{R}^{C}\rightarrow \mathbb{R}^{C_a},
\label{eq:down}
\end{equation}
where $\mathbf{U}\in\mathbb{R}^{B\times C_a\times T\times H\times W}$ is the bottleneck feature tensor. To apply temporal operators independently at each spatial location, we flatten spatial positions and obtain
\begin{equation}
\mathbf{U}\rightarrow \mathbf{U}_{f}\in\mathbb{R}^{(BHW)\times C_a\times T},
\label{eq:flatten}
\end{equation}
where the flattened index $m\in\{1,\dots,BHW\}$ enumerates spatial locations across the batch.

\paragraph{Formulation of temporal specialization.}
The novel design of DiA is an explicit decomposition of temporal evidence into two asymmetric pathways defined over $\mathbf{U}_{f}$, so that each spatial token can separately model past-conditioned and future-conditioned motion evolution. This is important in fine-grained action recognition because categories are often distinguished not only by \emph{what} motion occurs, but by \emph{how} it unfolds directionally over time.

The \emph{causal} direction aggregates only past-to-present evidence:
\begin{equation}
\mathbf{Y}^{\rightarrow} = f_{\mathrm{causal}}(\mathbf{U}_{f}),
\label{eq:causal}
\end{equation}
where $\mathbf{Y}^{\rightarrow}\in\mathbb{R}^{(BHW)\times C_a\times T}$ and $f_{\mathrm{causal}}(\cdot)$ is a depth-wise temporal convolution with kernel size $k$ and left replication padding. More explicitly, for flattened index $m$, channel $c\in\{1,\dots,C_a\}$, temporal index $t\in\{1,\dots,T\}$, and kernel offset $\delta\in\{0,\dots,k-1\}$,
\begin{equation}
\mathbf{Y}^{\rightarrow}_{m,c,t}
=
\sum_{\delta=0}^{k-1}
w^{\rightarrow}_{c,\delta}\,
\mathbf{U}^{\mathrm{pad}}_{f,m,c,t-\delta},
\end{equation}
where $w^{\rightarrow}_{c,\delta}$ denotes the causal filter weight for channel $c$ and offset $\delta$, and $\mathbf{U}^{\mathrm{pad}}_{f}$ denotes the left-padded sequence.

The \emph{anti-causal} direction aggregates only present-and-future evidence:
\begin{equation}
\mathbf{Y}^{\leftarrow} = f_{\mathrm{anti}}(\mathbf{U}_{f}),
\label{eq:anticausal}
\end{equation}
where $\mathbf{Y}^{\leftarrow}\in\mathbb{R}^{(BHW)\times C_a\times T}$ and $f_{\mathrm{anti}}(\cdot)$ is a depth-wise temporal convolution with right side replication padding:
\begin{equation}
\mathbf{Y}^{\leftarrow}_{m,c,t}
=
\sum_{\delta=0}^{k-1}
w^{\leftarrow}_{c,\delta}\,
\mathbf{U}^{\mathrm{pad}}_{f,m,c,t+\delta},
\end{equation}
where $w^{\leftarrow}_{c,\delta}$ denotes the anti-causal filter weight for channel $c$ and offset $\delta$.

This novel formulation imposes an explicit inductive bias toward temporal asymmetry and enables the adapter to preserve distinctions between preparation, execution, and completion phases that are typically collapsed by symmetric temporal updates.

\paragraph{Learnable directional fusion.}
Rather than enforcing equal contributions from both directions, we learn a global directional weighting. Let
\begin{equation}
\mathbf{a}=[a_{\rightarrow}, a_{\leftarrow}] \in \mathbb{R}^{2}
\end{equation}
be trainable direction logits, where $a_{\rightarrow}$ and $a_{\leftarrow}$ correspond to the causal and anti-causal directions, respectively. We normalize them using softmax:
\begin{equation}
[\lambda_{\rightarrow}, \lambda_{\leftarrow}]
=
\mathrm{Softmax}(\mathbf{a}),
\label{eq:softmax_weight}
\end{equation}
where $\lambda_{\rightarrow},\lambda_{\leftarrow}\in(0,1)$ and $\lambda_{\rightarrow}+\lambda_{\leftarrow}=1$. The fused directional response is then
\begin{equation}
\mathbf{Y}_{\mathrm{dir}}
=
\lambda_{\rightarrow}\mathbf{Y}^{\rightarrow}
+
\lambda_{\leftarrow}\mathbf{Y}^{\leftarrow},
\label{eq:direction_fusion}
\end{equation}
where $\mathbf{Y}_{\mathrm{dir}}\in\mathbb{R}^{(BHW)\times C_a\times T}$. This lets the model adaptively determine whether past-oriented or future-oriented evidence is more informative for the task.

\paragraph{Output update.}
The fused response $\mathbf{Y}_{\mathrm{dir}}$ is reshaped back to $\mathbb{R}^{B\times C_a\times T\times H\times W}$, passed through GELU, projected to the original channel dimension by $\phi_{\uparrow}$, and added residually to $\mathbf{V}$ to produce $\widehat{\mathbf{V}}$, which is then converted back to token form.

\paragraph{Identity-preserving initialization.}
The output projection $\phi_{\uparrow}$ is zero-initialized so that DiA starts from an identity residual mapping and learns task-specific temporal corrections progressively.

\paragraph{Compact functional form of DiA.}
The full DiA transformation can be written as
\begin{equation}
\widehat{\mathbf{X}}
=
[\mathbf{x}_{\mathrm{cls}};\,
\mathbf{X}_{p}
+
\phi_{\uparrow}
(
\sigma(
\mathcal{F}_{\mathrm{dir}}(
\phi_{\downarrow}(\mathbf{X}_{p})
)
)
)],
\label{eq:compact_dia}
\end{equation}
where $\mathcal{F}_{\mathrm{dir}}(\cdot)$ denotes the directional temporal operator defined by
\begin{equation}
\mathcal{F}_{\mathrm{dir}}(\cdot)
=
\lambda_{\rightarrow} f_{\mathrm{causal}}(\cdot)
+
\lambda_{\leftarrow} f_{\mathrm{anti}}(\cdot).
\end{equation}

DiA formulates temporal specialization into \textbf{asymmetric causal and anti-causal} decomposition with \textbf{learnable directional fusion}, enabling phase-sensitive fine-grained action refinement on frozen CLIP through parameter-efficient, token-local residual adaptation.

\begin{algorithm}[ht]
\caption{DiA: Directional Adapter (Forward pass of one DiA module)}
\label{alg:dia}
\begin{algorithmic}[1]
\REQUIRE Input token tensor $\mathbf{X}\in\mathbb{R}^{(HW+1)\times(BT)\times C}$, frozen CLIP backbone $f_{\theta}$, DiA parameters $\psi=\{\phi_{\downarrow},\phi_{\uparrow},f_{\mathrm{causal}},f_{\mathrm{anti}},\mathbf{a}\}$, number of frames $T$, temporal kernel size $k$
\ENSURE Refined token tensor $\widehat{\mathbf{X}}$

\STATE Split tokens: $\mathbf{X}\leftarrow[\mathbf{x}_{\mathrm{cls}};\mathbf{X}_{p}]$
\STATE Reshape patch tokens into video tensor $\mathbf{V}\in\mathbb{R}^{B\times C\times T\times H\times W}$
\STATE $\mathbf{U}\leftarrow \phi_{\downarrow}(\mathbf{V})$
\STATE Flatten spatial positions: $\mathbf{U}_{f}\in\mathbb{R}^{(BHW)\times C_a\times T}$
\STATE $\mathbf{Y}^{\rightarrow}\leftarrow f_{\mathrm{causal}}(\mathbf{U}_{f})$ \hfill $\triangleright$ past-to-present temporal modeling
\STATE $\mathbf{Y}^{\leftarrow}\leftarrow f_{\mathrm{anti}}(\mathbf{U}_{f})$ \hfill $\triangleright$ present-and-future temporal modeling
\STATE $[\lambda_{\rightarrow},\lambda_{\leftarrow}] \leftarrow \mathrm{Softmax}([a_{\rightarrow},a_{\leftarrow}])$
\STATE $\mathbf{Y}_{\mathrm{dir}} \leftarrow \lambda_{\rightarrow}\mathbf{Y}^{\rightarrow} + \lambda_{\leftarrow}\mathbf{Y}^{\leftarrow}$ \hfill $\triangleright$ directional fusion
\STATE Reshape $\mathbf{Y}_{\mathrm{dir}}$ back to $\mathbb{R}^{B\times C_a\times T\times H\times W}$
\STATE $\Delta \mathbf{V}\leftarrow \phi_{\uparrow}\big(\sigma(\mathbf{Y}_{\mathrm{dir}})\big)$
\STATE $\widehat{\mathbf{V}}\leftarrow \mathbf{V}+\Delta \mathbf{V}$
\STATE Convert $\widehat{\mathbf{V}}$ back to patch-token layout to obtain $\widehat{\mathbf{X}}_{p}$
\STATE $\widehat{\mathbf{X}}\leftarrow[\mathbf{x}_{\mathrm{cls}};\widehat{\mathbf{X}}_{p}]$
\RETURN $\widehat{\mathbf{X}}$
\end{algorithmic}
\end{algorithm}

Final classification follows the standard CLIP ViT pipeline. Frames are processed by the frozen CLIP transformer, with DiA inserted after the attention and MLP sub-blocks. DiA updates only the patch tokens, while subsequent self-attention propagates the refined information to the $[CLS]$ token. After the final LayerNorm, frame-wise $[CLS]$ features are temporally averaged and passed to a linear classifier for fine-grained action recognition.
\section{Results and discussion}
\label{sec:exp_results}

% =========================
% FineGym99 (Element-level action recognition on Gym99 subset)
% Source for accuracies: Table 3 in arXiv:2110.05853 (FineGym99 SOTA comparison)
% NOTE: Params/FLOPs are not reported in that source table → shown as “--” to avoid hallucination.
% =========================

% https://openaccess.thecvf.com/content/ICCV2021/papers/Chung_HAA500_Human-Centric_Atomic_Action_Dataset_With_Curated_Videos_ICCV_2021_paper.pdf
% https://github.com/kennymckormick/pyskl/blob/main/configs/posec3d/README.md

% StructViT-B-4-2 53.8 89.3 32 frames 51M params
% StructViT-B-4-1 54.2 89.5

% Our experiment is conducted on FineGym dataset [7],
% specifically Gym99 which contains annotation on 4 events,
% 15 sets, and 99 elements. At the point where we download
% the dataset, some of the videos are no longer available.

%%%%%%%%%%%%%%%%%%%%%%%%%%%Diving48%%%%%%%%%%%%%%%%%%%%%%%%%%%%%%%
\begin{table}[th]
\caption{\textbf{Results on Diving48.} DiA achieves the SoTA with substantially fewer tunable parameters. Parameter count $3.6$M is for Diving48 and changes slightly based the dataset. Total inference parameters: $85.8\text{M}$ (CLIP ViT-B/16) $+\,3.6\text{M}$ (DiA) $= 89.4\text{M}$.}
\small
\centering
\begin{tabular}{lcccc}
\toprule
\textbf{Method} & \textbf{Venue} &
\textbf{\begin{tabular}[c]{@{}c@{}}Tunable\\Params (M)\end{tabular}} &
\textbf{Frames} &
\textbf{Top-1} \\
\midrule
\multicolumn{4}{l}{\textit{Non-CLIP-based methods}} \\
\midrule
%TimeSformer-L \cite{bertasius2021space}          & 304.0 & 96                         & 81.0 \\
ORViT TimeSformer \cite{herzig2022object} & CVPR'22       & 160.0 & 32                         & 88.0 \\
TimeSformer \cite{bertasius2021space} &  ICML'21      & 121.4 & 16                         & 78.0 \\
BEVT \cite{wang2022bevt}    &     CVPR'22                & 88.1  & 16    & 86.7 \\
VideoSwin-B \cite{liu2022video} & CVPR'22                & 88.1  & 32   & 81.9 \\
SIFAR-B-14 \cite{fan2021can}   &  ICLR'22                & 87.0  & 16                         & 87.3 \\
SlowFast R101 \cite{feichtenhofer2019slowfast} & ICCV'19 & 62.8  & 64 + 16    & 77.6 \\
GC-TDN \cite{hao2022group}     &     CVPR'22             & 27.4  & 16                         & 87.6 \\
%PoseConvC3D \cite{duan2022revisiting}            & 2.0   & 48                         & 54.5 \\
TQN \cite{zhang2021temporal}    &      CVPR'21           & --    & all                        & 81.8 \\
CTM \cite{qian2024ctm}    &      CVIU'24           & 24.4    & -                        & 86.5 \\
ART \cite{sun2025toward}    &      TNNLS'25           & 365.0    & -                        & 87.9 \\

\midrule
\multicolumn{4}{l}{\textit{CLIP-based methods ViT-B/16}} \\
\midrule
BDC-CLIP \cite{long2025bdc}  &    ICML'25                & 125.4 & 16                         & 76.6 \\
EVL \cite{lin2022frozen} &     ECCV'22                    & 32.9  & 16                         & 51.42 \\
PeVL \cite{zhang2024pevl} &    CVPR'24         & 28.0  & 16                         & 84.5 \\
DiST \cite{qing2023disentangling} &   ICCV'23            & 19.0  & 16                         & 76.85 \\
AIM \cite{yang2023aim}     &     ICLR'23        & 11.0  & 32    & 88.9 \\
DualPath \cite{park2023dual}  &     CVPR'23              & 10.0  & 32    & 88.7 \\
ST Adapter \cite{pan2022st}   &    NeurIPS'22                  & 7.2   & 16   & 84.2 \\
\midrule
\rowcolor{green!10}
DiA (ours)              &        -                 & \textbf{3.6} & 32       & \textbf{92.8 $\pm$ 0.15} \\
\bottomrule
\end{tabular}
\label{tab:diving48_sota}
\end{table}

\subsection{Experimental details}
We evaluate DiA on challenging fine-grained action recognition benchmarks: Diving48 \cite{li2018diving48}, FineGym99/288 \cite{shao2020finegym}, HAA500 \cite{chung2021haa500}, ToyotaSmartHome \cite{das2019toyota}, and MPII Cooking 2 \cite{rohrbach15ijcv}, repeated with three random seeds. We use the standard splits for Diving48 , FineGym99/288, HAA500, and ToyotaSmartHome , reporting ToyotaSmartHome under the cross-subject (CS) protocol. For MPII Cooking 2, we use 67 classes with a 70/30 train/test split. Beyond fine-grained action recognition benchmarks, we also report results on two challenging datasets, HMDB51 \cite{kuehne2011hmdb} and Something-Something v2 \cite{goyal2017something}.

All experiments use CLIP ViT-B/16 as the visual backbone, with DiA inserted after each transformer block. We train for 50 epochs following the AIM \cite{yang2023aim} protocol and use 32 input frames by default. Unless otherwise specified, we use AdamW with cosine learning-rate decay (section \ref{sec:imple_details} in appendix). 

DiA uses CLIP ViT-B/16 ($C=768$, 12 layers), adapting all 12 transformer blocks. We use a bottleneck ratio of $\alpha=0.25$ ($C_a=192$) and directional depth-wise temporal convolutions with kernel size $k=5$ and replication padding. Each layer uses a learnable directional fusion pair initialized as $[0.5, 0.5]$ and normalized with softmax. Training uses 32-frame clips and AdamW with a learning rate of $0.0025$, weight decay of $0.05$, cosine learning-rate decay, and 2--3 warm-up epochs for 50 epochs. These settings are shared across datasets except for the classifier output dimension.

% More details in section \ref{sec:imple_details} in appendix.

%%%%%%%%%%%%%%%%%%%%%%%%%%%FineGym%%%%%%%%%%%%%%%%%%%%%%%%%%%%%%%

\begin{table*}[t]
\caption {Results on \textbf{FineGym99} and \textbf{FineGym288.} * means result produced.}
\centering
\small
\begin{tabular}{l c c c c c}
\toprule
\textbf{Method} & \textbf{Venue} & \textbf{Frames} & \textbf{\shortstack{Tunable\\Params (M)}} & \multicolumn{2}{c}{\textbf{Top-1}} \\
\cmidrule(lr){5-6}
 &  &  & & \textbf{FineGym99} & \textbf{FineGym288} \\
\midrule
\multicolumn{5}{l}{\textit{Non-CLIP-based methods}} \\
\midrule
TQN \cite{zhang2021temporal} & CVPR'21  & all & 78.0 & 93.8 & 89.6 \\
TSM \cite{lin2019tsm}   & ICCV'19 & 16+16 & 48.6 & 88.4 & 83.1 \\
TRNms \cite{zhou2018temporal} & ECCV'18 & 8+8 & 36.6 & 87.8 & 82.0 \\
SlowOnly \cite{feichtenhofer2019slowfast}  & ICCV'19 & 32 & 32.4 & 93.46 & 86.8 \\
I3D \cite{ao2017quo} & CVPR'17 & 16 & 28.0 & 75.6 & 77.3 \\
TSN \cite{wang2018temporal}     &  TPAMI'18 & 8 & 24.3 & 86.0 & 79.9 \\
ART \cite{sun2025toward}     &  TNNLS'25 & - & 365.0 & 94.7 & 90.2 \\

%3D VE \cite{chee2022combined} & 32 & 23.0 & 94.0 & - \\
%VT CE \cite{chee2022combined} & 32 & 16.0 & 94.6 & 90.1 \\
%PoseConv3D \cite{duan2022revisiting} & 48 & 2.0 & 93.2 & - \\
\midrule
\multicolumn{5}{l}{\textit{CLIP-based methods ViT-B/16}} \\
\midrule
PeVL*  \cite{zhang2024pevl} & CVPR'24 & 16 & 28.0 & 91.2 & 85.4 \\
AIM  \cite{yang2023aim}  & ICLR'23 & 32 & 11.0 & 94.9 & 89.1 \\
\midrule
\rowcolor{green!10}
DiA (ours) & - & 32 & 3.0 & \textbf{96.2 $\pm$ 0.13 } & \textbf{91.7 $\pm$ 0.10} \\

\bottomrule
\end{tabular}
\label{tab:finegym_sota}
\end{table*}

% =========================
% HAA500 (Atomic action classification)
% Source for accuracies: ICCV'21 HAA500 paper (openaccess.thecvf.com) Table snippet reporting RGB / Pose / RGB+Pose
% NOTE: This table reports accuracy (not Params/FLOPs) → shown as “--”.

% =========================

% =========================
% Toyota Smarthome (Video action classification)
% Source for accuracies: OpenCodePapers leaderboard page for Toyota Smarthome
% NOTE: Leaderboard reports CS/CV1/CV2 metrics; Params/FLOPs not provided → “--”.
% https://opencodepapers-b7572d.gitlab.io/benchmarks/action-classification-on-toyota-smarthome.html
% =========================

\begin{table*}[t]
\centering
\caption{Results on \textbf{HAA500} and \textbf{ToyotaSmartHome}.}
\label{tab:haa500_tsh_sota}

% -------- TOP ROW --------
\begin{minipage}[t]{0.48\textwidth}
\centering
\footnotesize
\setlength{\tabcolsep}{4pt}
\renewcommand{\arraystretch}{1.05}

\begin{tabular}{p{3.2cm} c c}
\toprule
\textbf{Method} & \textbf{\shortstack{Tunable\\Params (M)}} & \textbf{\shortstack{HAA500\\Top-1}} \\
\midrule
\multicolumn{3}{l}{\textit{Non-CLIP-based methods}} \\
\midrule
TPN \cite{yang2020temporal} & 33.0 & 50.5 \\
%I3D (RGB) & 28.0 & 33.53 \\
TSN \cite{wang2018temporal} & 24.3 & 55.3 \\
I3D \cite{ao2017quo} & 28.0 & 42.80 \\
SlowFast \cite{feichtenhofer2019slowfast} & 28.7 & 25.07 \\
\midrule
\multicolumn{3}{l}{\textit{CLIP-based methods ViT-B/16}} \\
\midrule
P2S + EVL \cite{li2024isolated} & 29.9 & 80.87 \\
EVL \cite{lin2022frozen} & 28.7 & 76.4 \\
PeVL*  \cite{zhang2024pevl} & 28.0 & 78.88 \\
AIM  \cite{yang2023aim}   & 11.0 & 80.2 \\
\midrule
\rowcolor{green!10}
DiA (ours) & \textbf{2.3} & \textbf{82.4 $\pm$ 0.08} \\
\bottomrule
\end{tabular}
\end{minipage}
\hfill
\begin{minipage}[t]{0.48\textwidth}
\centering
\footnotesize
\setlength{\tabcolsep}{4pt}
\renewcommand{\arraystretch}{1.05}
%\caption{Toyota Smarthome action classification results.}
\begin{tabular}{p{3.2cm} c c}
\toprule
\textbf{Method} & \textbf{\shortstack{Tunable\\Params (M)}} &  \textbf{\shortstack{ToyotaSH\\Top-1}} \\
\midrule
\multicolumn{3}{l}{\textit{Non-CLIP-based methods}} \\
\midrule
TimeSformer & 121.4 & 68.4 \\
$\pi$-ViT \cite{reilly2024just} & 121.4 & 72.9 \\
MotionFormer \cite{patrick2021keeping}& 109.1 & 65.8 \\
Video Swin \cite{liu2022video}& 88.1 & 69.8 \\
AssembleNet++ \cite{ryoo2020assemblenet}& 51.1 & 63.6 \\
MMNet \cite{bruce2022mmnet} & 34.2 & 70.1 \\
VPN++ \cite{das2021vpn}& 17.5 & 71.0 \\
PAAT \cite{reilly2023seeing} & -- & 72.5 \\
\midrule
\multicolumn{3}{l}{\textit{CLIP-based methods ViT-B/16}} \\
\midrule
PeVL* \cite{zhang2024pevl} & 28.0 & 69.3 \\
AIM \cite{yang2023aim}  & 11.0 & 84.3 \\

\midrule
\rowcolor{green!10}
DiA (ours) & \textbf{3.7} & \textbf{86.8 $\pm$ 0.12} \\
\bottomrule
\end{tabular}
\end{minipage}

\vspace{-2mm}
\end{table*}

% =========================
% HMDB and MPII Cooking

\begin{table*}[t]
\centering
\caption{Results on \textbf{HMDB51 } and \textbf{MPII Cooking 2 }.}
\label{tab:hmdb_mpii_sota}

% -------- LEFT TABLE (smaller) --------
\begin{minipage}[t]{0.42\textwidth}
\centering
\scriptsize
\setlength{\tabcolsep}{3pt}
\renewcommand{\arraystretch}{1.03}

\textbf{HMDB51}

\vspace{2pt}

\begin{tabular}{p{2.5cm} c c}
\toprule
\textbf{Method} & \textbf{\shortstack{Tunable\\Params (M)}} & \textbf{Top-1} \\
\midrule
\multicolumn{3}{l}{\textit{Non-CLIP-based methods}} \\
\midrule
SlowOnly-R101 \cite{duan2020omni} & 62.8 & 79.0 \\
I3D \cite{ao2017quo} & 28 & 74.8 \\
\midrule
\multicolumn{3}{l}{\textit{CLIP-based methods ViT-B/16}} \\
\midrule
Full-Tuning \cite{dosovitskiy2020image} & 86 & 59.4 \\
AIM \cite{yang2023aim} & 11.0 & 82.4 \\
DUALPATH \cite{park2023dual} & 10 & 75.6 \\
ST-Adapter \cite{pan2022st} & 7.20 & 77.7 \\
VPT \cite{jia2022visual} & 7 & 62.4 \\
VideoPrompt \cite{ju2022prompting} & 6 & 66.4 \\
\midrule
\rowcolor{green!10}
DiA (ours) & \textbf{3.6} & \textbf{85.6 $\pm$ 0.10} \\
\bottomrule
\end{tabular}
\end{minipage}
\hfill
% -------- RIGHT TABLE --------
\begin{minipage}[t]{0.54\textwidth}
\centering
\footnotesize
\setlength{\tabcolsep}{4pt}
\renewcommand{\arraystretch}{1.05}

\textbf{MPII Cooking 2}

\vspace{2pt}

\begin{tabular}{p{3.2cm} c c}
\toprule
\textbf{Method} & \textbf{\shortstack{Tunable\\Params (M)}} & \textbf{Top-1} \\
\midrule
\multicolumn{3}{l}{\textit{CLIP-based methods ViT-B/16}} \\
\midrule
AIM \cite{yang2023aim} & 11.0 & 76.1 \\
DualPath \cite{park2023dual} & 10.0 & 74.8 \\
ST \cite{pan2022st} & 7.20 & 75.4 \\
\midrule
\rowcolor{green!10}
DiA (ours) & \textbf{3.7} & \textbf{78.7 $\pm$ 0.15} \\
\bottomrule
\end{tabular}
\end{minipage}
\vspace{-3mm}
\end{table*}
%%%%%%%%%%%%%%%%%%%%%%%

\subsection{Consistent state-of-the-art performance across benchmarks}
\label{sec:sota_results}
DiA delivers consistent state-of-the-art performance among single-stream methods using only the video modality (RGB) across fine-grained actions in sports, atomic human actions, daily activities, cooking actions, and general action-recognition benchmarks. 

On Diving48, DiA achieves 92.8\% Top-1 accuracy (Table~\ref{tab:diving48_sota}) with only 3.6M tunable parameters, improving over AIM from 88.9\% while reducing the tunable parameters to 1/4th. The trend remains consistent on FineGym99/288 , where DiA obtains 96.2\% and 91.7\% Top-1 accuracy (Table~\ref{tab:finegym_sota}), compared with AIM at 94.9\% and 89.1\% under the same 32-frame setting. These substantial gains reiterate that Diving48 and FineGym require resolving phase-sensitive temporal differences, such as rotation count, transition order, and execution dynamics, supporting DiA’s direction-aware temporal modeling over symmetric aggregation.

DiA also improves over AIM on HAA500 and ToyotaSmartHome (Table~\ref{tab:haa500_tsh_sota}), as well as MPII Cooking 2 and HMDB51 (Table~\ref{tab:hmdb_mpii_sota}), reaching 82.4\%, 86.8\%, 78.7\%, and 85.6\% Top-1 accuracy, respectively. This cross-domain consistency indicates that directional temporal adaptation is not limited to sports actions, but also benefits daily activities, procedural hand-object interactions, and broader action-recognition settings. Additional results on SSv2 \cite{goyal2017something} in appendix section \ref{sec:ssv2}.

Using only the RGB video stream: no pose, language, or auxiliary cues, DiA matches or surpasses methods using extra modalities. With matched backbones, it outperforms multimodal PeVL (video + pose + text): at ViT-L/14 it achieves 93.8 vs. 92.5 on Diving48 and 92.9 vs. 91.8 on FineGym288, using 12.7M vs. 109M tunable parameters. Under the same RGB-only setting, video-only PeVL scores only 84.5 on Diving48 versus DiA's 92.8, indicating much of PeVL's gain comes from auxiliary streams rather than stronger temporal modeling. DiA therefore remains a parameter- and compute-efficient temporal adapter that achieves state-of-the-art performance using a single input stream. 

Few-shot results on UCF101 \cite{soomro2012ucf101} and HMDB51  \cite{kuehne2011hmdb} are provided in Appendix~\ref{app:fewshot_ucf_hmdb}.

%\vspace{-3mm}

\subsection{Achieving the accuracy--compute efficiency--unimodality objective}
\label{sec:fullsup_efficiency_dia_aim}

The accuracy--compute efficiency--unimodality objective asks whether temporal specialization can improve recognition while remaining compact, fast, and single-stream. We assess this trade-off under the same full-supervised Diving48 protocol on a single NVIDIA A100-SXM4-40GB GPU with FP16 inference. FLOPs are computed with \texttt{fvcore}, and latency is measured with CUDA events using batches of 8 clips, each with shape $[3,32,224,224]$.

Table~\ref{tab:fullsup_efficiency_dia_aim} shows that DiA improves accuracy and efficiency simultaneously. Accuracy increases from 88.9\% to 92.8\%, while per-clip computation decreases from \SI{812.06}{GFLOP} to \SI{558.37}{GFLOP}, giving a 31.24\% FLOP reduction. The same reduction appears at the batch level, showing that the gain is not an artifact of per-sample reporting. The runtime measurements confirm the practical impact. DiA increases throughput from \SI{28.15}{clips/s} to \SI{61.20}{clips/s} and reduces p50 batch latency, i.e., the median wall-clock time for one forward pass over a batch of 8 clips, from \SI{284.14}{ms} to \SI{130.73}{ms}. It also improves realized compute from \SI{22.86}{TFLOP/s} to \SI{34.17}{TFLOP/s}, indicating better use of the same GPU.
These results show that DiA achieves the intended objective: stronger recognition with lower FLOPs, lower latency, higher throughput, and higher realized compute, while retaining compact adapter-based training and a single-stream input setting.

Additionally, DiA consistently outperforms AIM across adaptation depths and frame settings while using substantially fewer tunable parameters, with performance saturating around 10 blocks and improving with increased temporal context (refer section \ref{sec:block_frame_ablation} in appendix).

\begin{table}[h!]
\vspace{-4mm}
\centering
\caption{\textbf{DiA compute-efficiency analysis on Diving48}. $\downarrow$ indicates lower is better, $\uparrow$ indicates higher is better.}
\label{tab:fullsup_efficiency_dia_aim}
\scriptsize
\setlength{\tabcolsep}{4pt}
\renewcommand{\arraystretch}{1.12}

\resizebox{\linewidth}{!}{%
\begin{tabular}{lccccccc}
\toprule
Method
& \shortstack{Top-1\\Acc. $\uparrow$}
& \shortstack{GFLOPs\\per clip $\downarrow$}
& \shortstack{GFLOPs per batch\\($B{=}8$) $\downarrow$}
& \shortstack{Throughput\\(clips/s) $\uparrow$}
& \shortstack{Effective compute\\(TFLOP/s) $\uparrow$}
& \shortstack{p50 latency\\(ms) $\downarrow$} \\
\midrule
AIM
& 88.9
& 812.06
& 6496.49
& 28.15
& 22.86
& 284.14 \\

DiA (ours)
& \textbf{92.8}
& \textbf{558.37}
& \textbf{4466.99}
& \textbf{61.20}
& \textbf{34.17}
& \textbf{130.73} \\
\midrule
\rowcolor{green!10}
Gain (DiA vs. AIM)
& $\uparrow 3.9$
& $\downarrow 31.24\%$
& $\downarrow 31.24\%$
& $\uparrow 2.17\times$
& $\uparrow 1.49\times$
& $\downarrow 53.98\%$ \\
\bottomrule
\end{tabular}%
}
\vspace{-3mm}
\end{table}

\subsection{How asymmetric directionality benefits DiA?}
\label{sec:why_dia}

\begin{table}[t]
\centering

\begin{minipage}[t]{0.38\linewidth}
\centering
\caption{Component ablation study on Diving48.}
\label{tab:component_ablation}
\footnotesize
\setlength{\tabcolsep}{5pt}
\renewcommand{\arraystretch}{1.1}
\begin{tabular}{l c}
\toprule
\textbf{Component} & \textbf{Top-1} \\
\midrule
With Causal & 91.3 \\
With Anti-causal & 91.5 \\
\midrule
Causal + Anti-causal (DiA) & 92.8 \\
\bottomrule
\end{tabular}
\end{minipage}
\hfill
\begin{minipage}[t]{0.58\linewidth}
\centering
\caption{Qualitative stats on Diving48 subset (\(N=120\)).}
\label{tab:qual_global}
\footnotesize
\setlength{\tabcolsep}{4pt}
\renewcommand{\arraystretch}{1.1}
\begin{tabular}{l c}
\toprule
\textbf{Metric} & \textbf{Value} \\
\midrule
Causal activity ratio & 1.0000 \\
Anti-causal activity ratio & 1.0000 \\
Mean \(\mathrm{corr}(E_c,E_a)\) & -0.1884 \\
Mean \(\mathrm{corr}(E_c,\mathrm{reverse}(E_a))\) & -0.1705 \\
Dominance mean \(\mathbb{E}[\bar D]\) & 0.0278 \\
Dominance std \(\mathrm{Std}(\bar D)\) & 0.0193 \\
Balanced-clip ratio \((|\bar D|\le 0.05)\) & 0.8833 \\
\bottomrule
\end{tabular}
\end{minipage}
\vspace{-4mm}
\end{table}

We evaluate DiA through component ablation and directional behavior. The ablation tests whether causal and anti-causal directions provide complementary temporal evidence, while the qualitative analysis verifies that both directions remain active and non-redundant after training.  As shown in Table~\ref{tab:component_ablation}, the full causal+anti-causal DiA achieves 92.8\% Top-1 accuracy, outperforming causal-only (91.3\%) and anti-causal-only (91.5\%) variants, indicating that the two directions capture complementary temporal evidence.

%%%%%%%%%%%%%%%%%%%%%%%%%%%%%%%%%%%

\subsubsection{Qualitative evidence for causal and anti-causal modeling}

\begin{figure}[h]
\centering
\includegraphics[width=\linewidth]{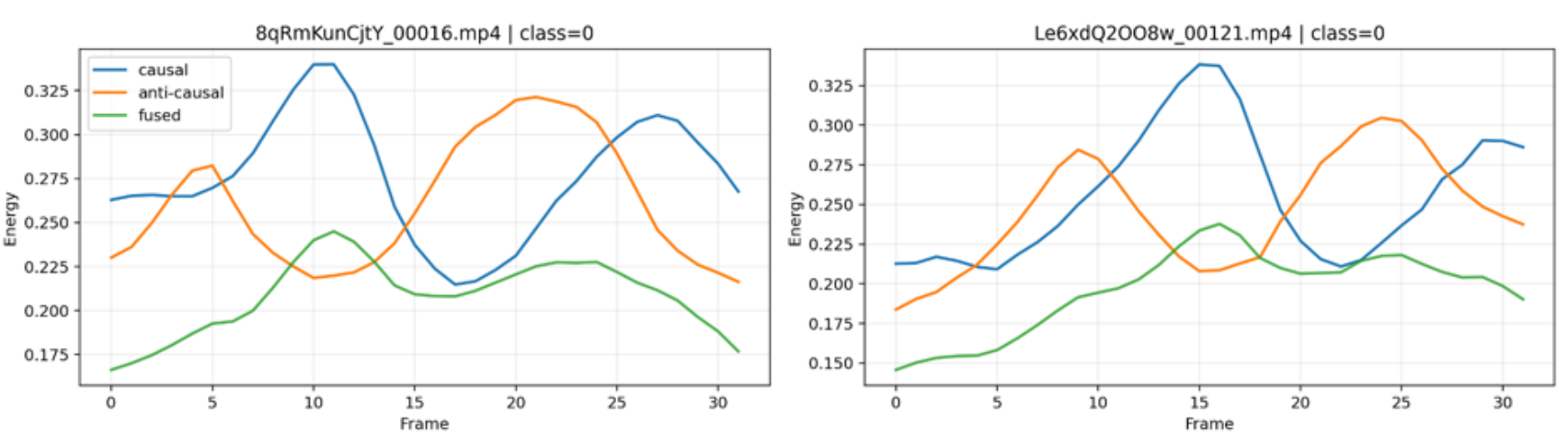}
\caption{
Representative direction-energy trajectories on Diving48 test clips. Causal and anti-causal responses peak at different frames, indicating that the two directions capture different temporal evidence, while the fused response combines them over the clip.}
\label{fig:qual_panel}
\vspace{-4mm}
\end{figure}

\begin{figure}[h]
\centering
\includegraphics[width=0.49\linewidth]{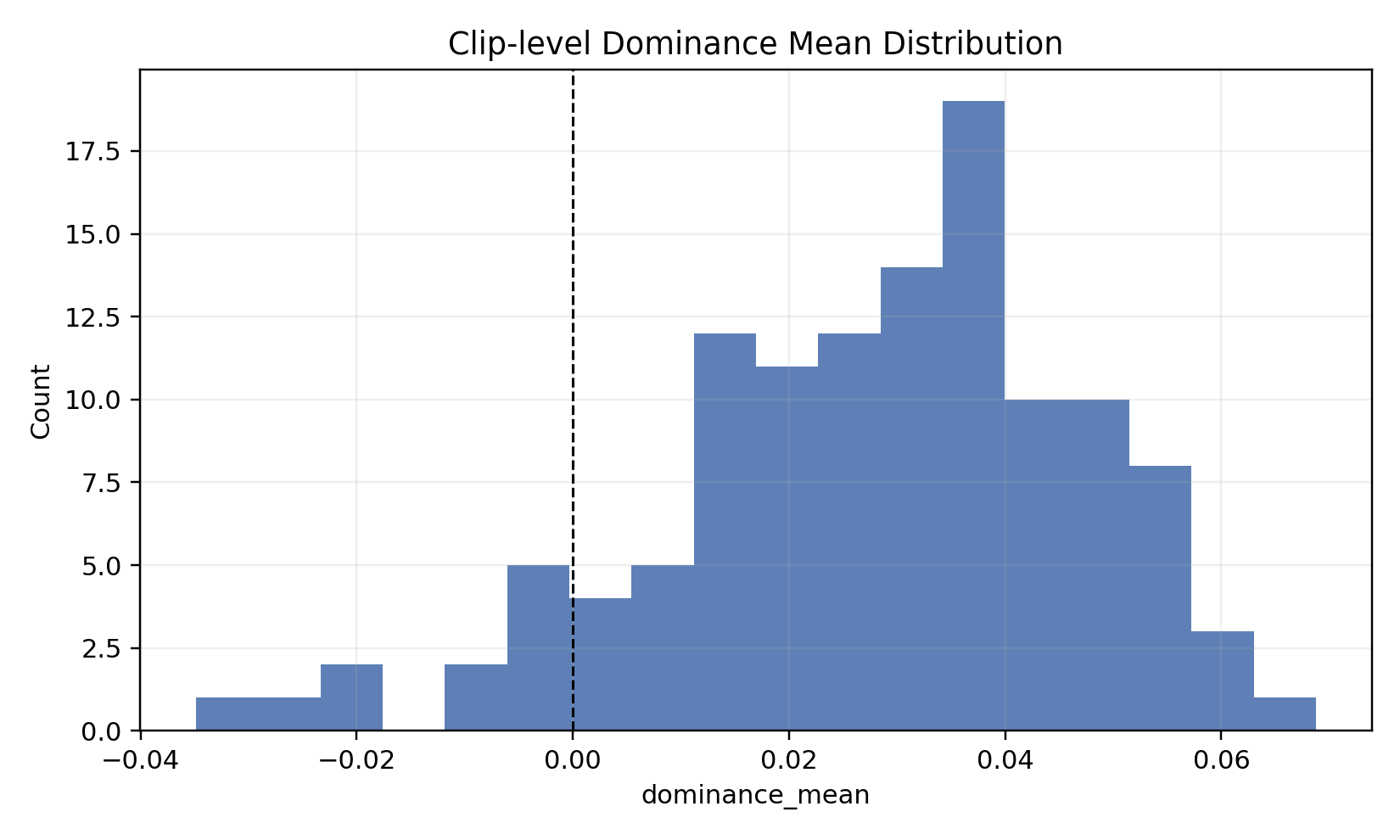}
\includegraphics[width=0.49\linewidth]{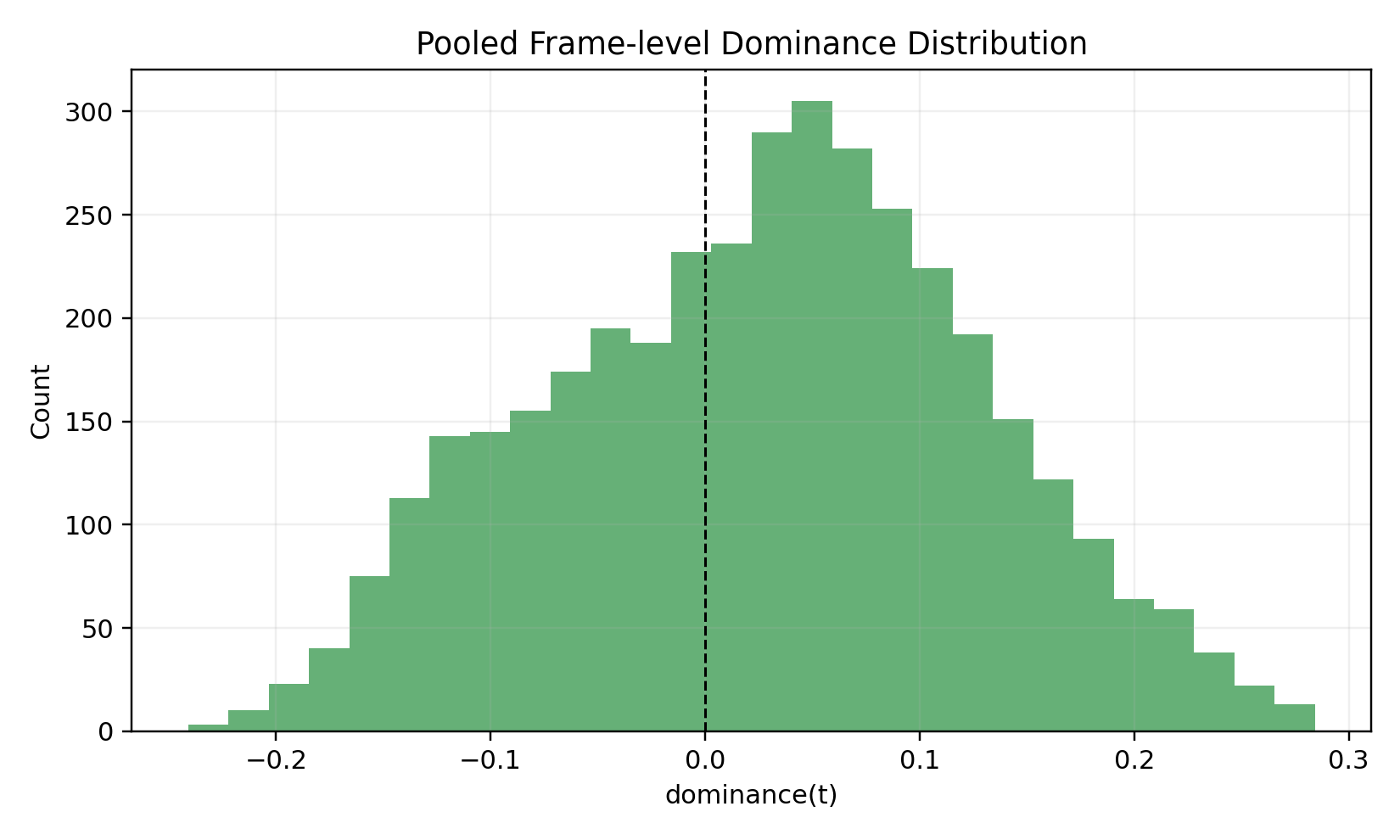}
\caption{
Directional dominance statistics on Diving48. Left: the clip-level mean dominance \(\bar D\) is concentrated near zero, showing balanced causal and anti-causal use over complete clips. Right: the frame-level dominance \(D(t)\) is more spread out, showing that one direction can dominate locally at particular action phases.}
\label{fig:qual_hist}
\vspace{-4mm}
\end{figure}

We further analyze whether the gain of DiA comes from complementary directional evidence rather than additional capacity alone. Using a held-out Diving48 test subset with \(N=120\) clips, we record causal, anti-causal, and fused activations across adapter layers, spatial tokens, channels, and frames. For each direction, we compute the frame-wise energy
\begin{equation}
E_b(t)=\mathbb{E}_{\ell,s,ch}\left[\left|z_{b,\ell,s,ch}(t)\right|\right],
\end{equation}
where \(b\in\{c,a,f\}\) denotes causal, anti-causal, and fused responses, and \(z_{b,\ell,s,ch}(t)\) is the activation at frame \(t\), layer \(\ell\), spatial token \(s\), and channel \(ch\). This energy measures response magnitude over time; it is not a confidence score. We measure directional dominance as
\begin{equation}
D(t)=\frac{E_c(t)-E_a(t)}{E_c(t)+E_a(t)+\epsilon}.
\end{equation}
Here, \(D(t)>0\) indicates stronger causal response, \(D(t)<0\) indicates stronger anti-causal response, and \(D(t)\approx0\) indicates balanced use of both directions.

Table~\ref{tab:qual_global} reports three diagnostics. Activity ratios test whether each direction remains active. Correlations between \(E_c\) and \(E_a\) test whether the two directions produce redundant temporal profiles. Dominance statistics test whether DiA collapses toward one direction. Both activity ratios are 1.0000, showing no directional collapse. The negative correlations, including \(\mathrm{corr}(E_c,\mathrm{reverse}(E_a))\), indicate that the anti-causal response is not a delayed or reversed copy of the causal response. The dominance statistics show near-balanced use, with \(\mathbb{E}[\bar D]=0.0278\) and 88.33\% of clips satisfying \(|\bar D|\le0.05\). These results support complementary directional modeling rather than a capacity-only explanation.

Fig.~\ref{fig:qual_panel} visualizes the temporal responses for representative clips. The causal and anti-causal energies often peak at different frames, showing that the two directions respond to different parts of the action rather than producing the same temporal profile. The fused response combines these directional responses over time. Fig.~\ref{fig:qual_hist} summarizes this pattern over the test subset. The clip-level dominance is concentrated near zero (shown in blue on left , Fig.~\ref{fig:qual_hist}), showing that DiA uses both directions over a full clip. In contrast, the frame-level dominance is more spread out (shown in green on right , Fig.~\ref{fig:qual_hist}), showing that causal or anti-causal evidence can be stronger at specific frames.

%Finally, we inspect the learned fusion weights \(\lambda_{\rightarrow}\) and \(\lambda_{\leftarrow}\) from Eq.~\eqref{eq:direction_fusion}. These weights are computed by the softmax over the trainable direction logits and determine how much the causal and anti-causal directions contribute in each adapter layer. Across the 10 adapted layers, the mean causal and anti-causal weights are 0.5030 and 0.4970, with a mean absolute gap of 0.0342. Thus, the fusion does not collapse to one direction. Together with the ablation and dominance analysis, this shows that DiA uses both past-conditioned and future-conditioned evidence for fine-grained recognition.
We analyze the learned fusion weights $\lambda_{\rightarrow}$ and $\lambda_{\leftarrow}$, Eq.~\eqref{eq:direction_fusion}, obtained by softmax over trainable direction logits. Across 10 layers, their means are 0.5030 and 0.4970, with a small mean gap of 0.0342, indicating no collapse to either direction. Together with ablation and dominance results, this shows that DiA uses both past- and future-conditioned evidence for fine-grained action recognition.

Further additional examples and complementary-branch analysis (per-layer, per-clip, per-class) are given in Appendix~\ref{sec:additional_quali}.
\section{Conclusion and limitations}
\label{sec:conclusion}
The central finding is that fine-grained action recognition benefits from directional temporal specialization under a tight efficiency budget. The gains over causal-only and anti-causal-only variants show that past- and future-conditioned cues are complementary, while the cross-dataset improvements show that this benefit is not obtained by full model adaptation or auxiliary modalities. DiA therefore demonstrates that a frozen visual backbone can be effectively specialized for phase- and order-sensitive actions with only a compact set of tunable parameters and lower inference cost.
Two directions remain open for the future work which are currently not addressed. First, studying longer-range fine-grained action structure will require curating long-duration fine-grained action benchmarks, as existing datasets contain relatively short clips and may not capture extended temporal dependencies. Second, DiA uses both causal and anti-causal directions within the observed clip, which is well suited to offline recognition but leaves online recognition as a natural extension, where future frames are not available at prediction time.

\bibliographystyle{plainnat}
\bibliography{example_paper}

\newpage
% \input{checklist}

% \newpage
\appendix
\appendix

\section{Analysis of Video-Only and Multimodal Recognition}
\label{sec:multimodal_analysis}

To isolate the contribution of the video modality, we compare DiA with PeVL~\cite{zhang2024pevl} under video-only and multimodal settings. PeVL combines video, pose, and text, whereas DiA uses only the video stream. Since the official PeVL code is not publicly available, we re-implemented its video-only setting. On Diving48, our reproduced result matches the video-only accuracy reported by PeVL (84.5\%); for FineGym99/288, where video-only results are not reported, our implementation obtains 91.2\% and 85.4\%, respectively.

\begin{table}[h]
\centering
\small
\setlength{\tabcolsep}{5pt}
\renewcommand{\arraystretch}{1.08}
\caption{\textbf{Video-only and multimodal comparison with PeVL.}
Top-1 accuracy (\%) is reported on Diving48 and FineGym99/288.
``Reported'' denotes results from the original PeVL paper \cite{zhang2024pevl}; the remaining PeVL results are reproduced using our implementation. DiA uses only the video modality.}
\label{tab:pevl_modality_comparison}
\begin{tabular}{lcccc}
\toprule
\textbf{Method} &
\textbf{\shortstack{Tunable\\Params (M)}} &
\textbf{Diving48} &
\textbf{FineGym99} &
\textbf{FineGym288} \\
\midrule

\multicolumn{5}{l}{\textit{ViT-B/16 backbone}} \\
\midrule
PeVL (video + pose + text), reported
    & 42.0 & 91.9 & \textbf{96.5} & 90.5 \\
PeVL (video)
    & 28.0 & 84.5 & 91.2 & 85.4 \\
DiA (video)
    & \textbf{3.7} & \textbf{92.8} & 96.2 & \textbf{91.7} \\

\midrule
\multicolumn{5}{l}{\textit{ViT-L/14 backbone}} \\
\midrule
PeVL (video + pose + text), reported
    & 109.0 & 92.5 & \textbf{97.0} & 91.8 \\
DiA (video)
    & \textbf{12.7} & \textbf{93.8} & -- & \textbf{92.9} \\

\bottomrule
\end{tabular}
\end{table}

Table~\ref{tab:pevl_modality_comparison} shows a clear modality--parameter trade-off. With ViT-B/16, removing pose and text from PeVL reduces accuracy from 91.9\% to 84.5\% on Diving48, indicating a substantial contribution from its additional modalities. In contrast, DiA reaches 92.8\% using only video and 3.7M tunable parameters. The trend remains at the larger ViT-L/14 backbone: DiA achieves 93.8\% versus 92.5\% on Diving48 and 92.9\% versus 91.8\% on FineGym288, while using 12.7M rather than 109M tunable parameters. Thus, DiA remains competitive with or exceeds the multimodal PeVL results under matched backbones, while operating with a single video input stream and substantially fewer tunable parameters.

\section{Additional results on Something-Something v2}
\label{sec:ssv2}
On SSv2 (Table \ref{ref:ssv2}), methods with the highest accuracy rely on large-capacity models, e.g., MViTv2-L (213M, 73.3\%), ViViT-L (311M, 65.4\%), and MTV-B (310M, 67.6\%), indicating that strong performance is often achieved through substantial parameter scaling. In contrast, parameter-efficient methods operate in a much smaller regime: AIM (14M, 68.1\%) and ST (7.2M, 69.3\%). DiA uses only 3.0M tunable parameters and achieves 69.2\% top-1 accuracy, remaining competitive within this low-parameter regime. 

This highlights a clear trade-off: while larger models achieve higher absolute accuracy, they require one to two orders of magnitude more tunable parameters. DiA prioritizes parameter and compute efficiency, achieving comparable performance to existing efficient methods with significantly fewer parameters. The gap to large-scale models reflects the increased temporal complexity of SSv2, where higher capacity can be beneficial, but DiA maintains a favorable accuracy–efficiency balance under strict parameter constraints.

\begin{table}[h]
\centering
\footnotesize
\setlength{\tabcolsep}{4pt}
\renewcommand{\arraystretch}{1.05}
\caption{Comparison on \textbf{Something-Something V2 (SSv2)}. We report Top-1 accuracy on a challenging temporal reasoning benchmark. DiA maintains competitive performance while using a compact set of tunable parameters.}
\label{ref:ssv2}
\begin{tabular}{p{3.5cm} c c}
\toprule
\textbf{Method} & \textbf{\shortstack{Tunable\\Params (M)}} & \textbf{Top-1} \\
\midrule
\multicolumn{3}{l}{\textit{Non-CLIP-based methods}} \\
\midrule
ViViT-L/16$\times$2 \cite{arnab2021vivit} & 311 & 65.4 \\
MTV-B \cite{yan2022multiview} & 310 & 67.6 \\
MViTv2-L (312$\uparrow$) \cite{li2022mvitv2} & 213 & \textbf{73.3} \\
TimeSformer-L \cite{bertasius2021space} & 121 & 62.4 \\
VideoSwin-B \cite{liu2022video} & 89 & 69.6 \\
MViTv2-B \cite{li2022mvitv2}& 51 & 70.5 \\
UniFormer-B \cite{li2022uniformer} & 50 & 71.2 \\
MViT-B \cite{fan2021multiscale}& 37 & 67.1 \\
CTM \cite{qian2024ctm}& 24.4 & 65.1 \\
Omnivore \cite{girdhar2022omnivore}& - & 71.4 \\
CoVeR \cite{zhang2021co}& - & 70.9 \\

\midrule
\multicolumn{3}{l}{\textit{CLIP-based methods}} \\
\midrule
EVL \cite{lin2022frozen} & 86 & 62.4 \\
AIM \cite{yang2023aim} & 14 & 68.1 \\
ST \cite{pan2022st} & 7.2 & 69.3 \\
\midrule
DiA (ours) & \textbf{3.0} & 69.2 \\
\bottomrule
\end{tabular}
\end{table}

\section{Ablation on number of blocks and frames}
\label{sec:block_frame_ablation}

We ablate the number of adapted blocks and input frames to evaluate whether DiA remains effective under different capacity and temporal-context budgets.

Table~\ref{tab:block_ablation} shows that DiA achieves strong performance even with limited adaptation depth. With only 6 adapted blocks, DiA already outperforms AIM while using substantially fewer tunable parameters. Performance improves further when increasing to 10 blocks, but saturates at 12 blocks. This indicates that DiA does not require dense adaptation across the full backbone; most of the benefit comes from a compact number of directionally specialized temporal updates. We therefore use 10 blocks as the best accuracy--parameter trade-off.

\begin{table*}[h]
\centering

\begin{minipage}[t]{0.48\textwidth}
\centering
\caption{Impact of adaptation depth on Diving48. DiA achieves higher accuracy than AIM with fewer tunable parameters across different numbers of adapted CLIP blocks.}
\label{tab:block_ablation}
\footnotesize
\setlength{\tabcolsep}{5pt}
\renewcommand{\arraystretch}{1.1}
\begin{tabular}{l c c c}
\toprule
\textbf{Method} & \textbf{\# Blocks} &  \textbf{\shortstack{Tunable\\Params (M)}} &\textbf{ Top-1} \\
\midrule
AIM & 6 & 5.3 & 87.3 \\
DiA & 6 &\textbf{ 1.8} & \textbf{90.6} \\ [4pt]

AIM & 10 & 8.9 & 87.7 \\
DiA & 10 & \textbf{3.0} & \textbf{92.7} \\[4pt]

AIM & 12 & 11.0 & 88.9 \\
DiA & 12 & \textbf{3.7} & \textbf{92.8} \\
\bottomrule
\end{tabular}
\end{minipage}
\hfill
% -------- RIGHT TABLE --------
\begin{minipage}[t]{0.48\textwidth}
\centering
\caption{Impact of temporal context on Diving48. DiA consistently outperforms AIM across frame budgets, showing that directional temporal adaptation uses available temporal evidence more effectively.}
\label{tab:frame_ablation}
\footnotesize
\setlength{\tabcolsep}{5pt}
\renewcommand{\arraystretch}{1.1}
\begin{tabular}{l c c}
\toprule
\textbf{Method} & \textbf{\# Frames} & \textbf{Top-1} \\
\midrule
AIM & 8  & 81.8 \\
DiA & 8 & \textbf{83.0} \\ [4pt]

AIM & 16 & 84.7 \\
DiA & 16 & \textbf{90.5} \\ [4pt]

AIM & 32 & 88.9 \\
DiA & 32 & \textbf{92.7} \\ [4pt]

\bottomrule
\end{tabular}
\end{minipage}

\end{table*}

Table~\ref{tab:frame_ablation} shows that DiA benefits from additional temporal context. The largest improvement appears when increasing from 8 to 16 frames, suggesting that directional temporal modeling needs sufficient temporal evidence to capture action phases. The gain from 16 to 32 frames is smaller but remains positive, indicating that longer clips still provide useful phase-completion cues. DiA also outperforms AIM at every frame setting, showing that the causal/anti-causal design uses available frames more effectively rather than simply relying on more input.

Overall, these ablations show that DiA is robust across both adaptation depth and temporal input length, while preserving the paper's accuracy--efficiency objective.

\section{Few-shot comparison on UCF101 and HMDB51}
\label{app:fewshot_ucf_hmdb}

To assess whether DiA remains effective when labelled data is scarce, we
evaluate it in the $K$-shot setting on UCF101~\cite{soomro2012ucf101} and
HMDB51~\cite{kuehne2011hmdb}, with $K \in \{2, 4\}$ training videos per class.
% TODO: state the protocol (e.g., which splits / how the K samples are drawn,
% number of runs, frames, epochs), or cite the protocol you follow.
We compare with Storyboard~\cite{liu2026storyboard} under the same setting.

\begin{table}[h]
\centering
\caption{Few-shot Top-1 accuracy on UCF101 and HMDB51 with $K$ training videos
per class.}
\label{tab:app_fewshot}
\small
\begin{tabular}{lcccc}
\toprule
\multirow{2}{*}{Method} & \multicolumn{2}{c}{UCF101} & \multicolumn{2}{c}{HMDB51} \\
\cmidrule(lr){2-3} \cmidrule(lr){4-5}
 & $K=2$ & $K=4$ & $K=2$ & $K=4$ \\
\midrule
Storyboard~\cite{liu2026storyboard} & 92.1 & 93.0 & 64.0 & 66.1 \\
DiA (ours) & \textbf{93.7} & \textbf{96.9} & \textbf{77.1} & \textbf{85.5} \\
\bottomrule
\end{tabular}
\end{table}

As shown in Table~\ref{tab:app_fewshot}, DiA outperforms Storyboard at both
shot counts on both datasets. The gains on UCF101 are 1.6 and 3.9 points. The
gains on HMDB51 are larger, at 13.1 and 19.4 points. These results indicate
that the compact directional adapter does not depend on large amounts of
labelled data and transfers well to the low-data regime.

\section{Implementation details}
\label{sec:imple_details}
We use CLIP ViT-B/16 as the visual backbone and train with a batch size of 8. For Diving48, FineGym99, HMDB51, HAA500, ToyotaSH and MPII Cooking 2, we sample 32 input frames with segment-based sampling, resize the short side to 256, and apply 224$\times$224 crops. For SSv2, we use 16 input frames with the same sampling and crop pipeline. Optimization follows AdamW with a default learning rate of 0.0025 and cosine learning-rate decay, warmup for 2--3 epochs, and weight decay in the range 0.03--0.05. We train for 50 epochs on Diving48, MPII Cooking 2, FineGym99, HAA500, ToyotaSH and HMDB51, and 45 epochs on SSv2. We follow standard video data processing, including random cropping, resizing, and horizontal flipping for augmentation, along with multi-threaded video decoding.

\section{Qualitative evidence for causal and anti-causal modeling: additional examples and complementary branches analysis}
\label{sec:additional_quali}

\begin{figure}[h]
\centering
\includegraphics[width=0.8\linewidth]{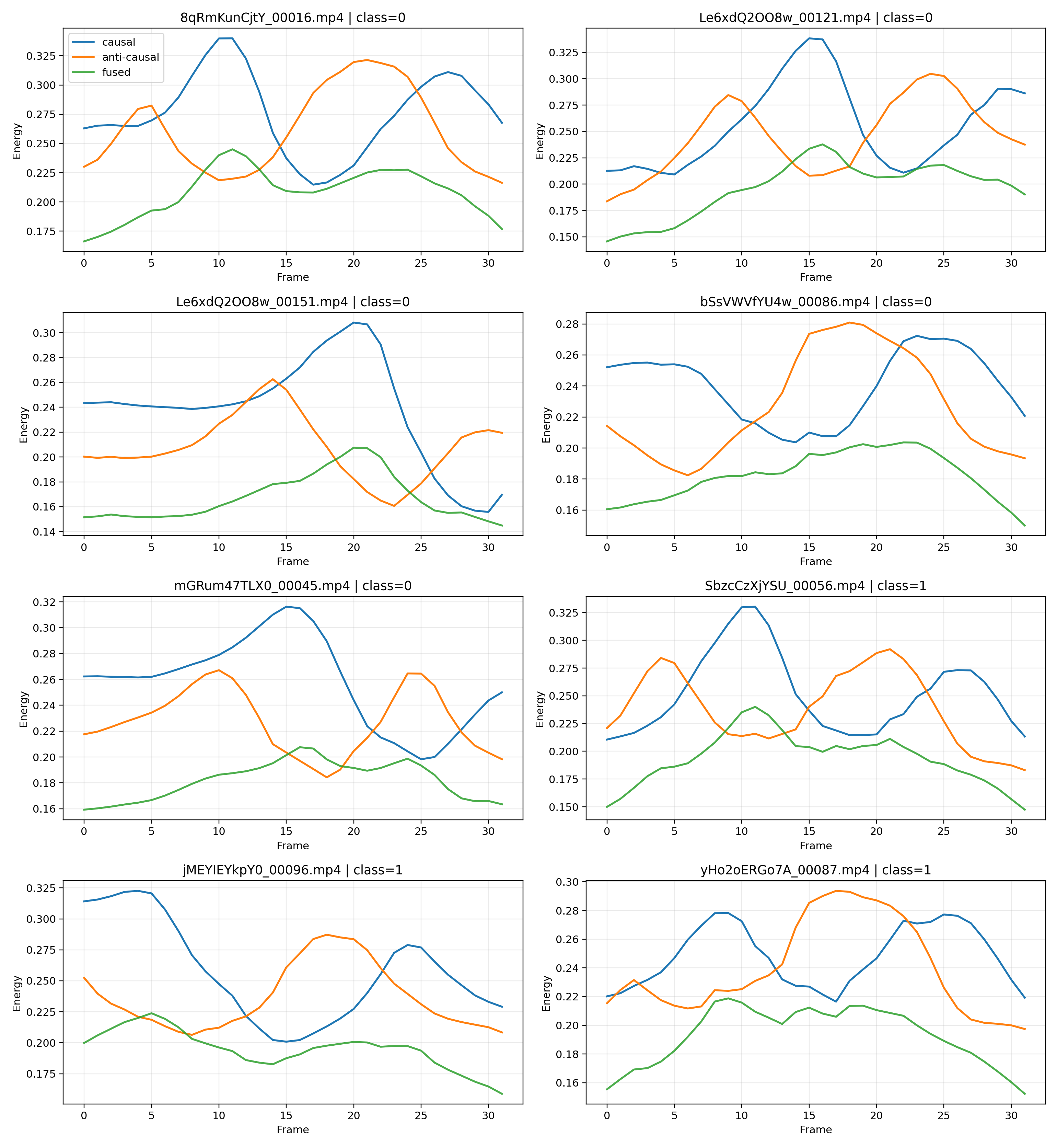}
\caption{
Representative direction-energy trajectories on Diving48 test clips. Each plot shows how causal, anti-causal, and fused responses vary over frames. Peaks at different temporal locations indicate that the two directions respond to different action phases, while the fused response combines their evidence over the clip.
}
\label{fig:qual_panel_extra}
\end{figure}

\subsection{Complementary branches analysis}
\label{app:complementary}

We extend the analysis of Sec.~4.4.1 on the same Diving48 test subset
($N = 120$ clips, 20 classes), using the energy and dominance measures of
Eqs.~(16)--(17). The component ablation (Table~6) already shows that each
direction is necessary: causal-only and anti-causal-only variants reach 91.3\%
and 91.5\%, respectively, compared with 92.8\% for the fused model. Below, we
report per-layer, per-clip, and per-class statistics of the trained model.

Table~\ref{tab:app_fusion_weights} shows the learned fusion weights
(Eq.~(12)) for each adapted block. The split remains close to $50/50$ at every
depth (mean $0.503/0.497$), with a maximum imbalance of $0.139$ at block~3.
Although the softmax could concentrate the weight on one direction, both
directions are retained throughout the network.

\begin{table}[h]
\centering
\caption{Learned directional fusion weights per adapted block on Diving48.
The larger weight in each block is shown in bold.}
\label{tab:app_fusion_weights}
\small
\setlength{\tabcolsep}{4.5pt}
\begin{tabular}{lcccccccccc}
\toprule
Block & 2 & 3 & 4 & 5 & 6 & 7 & 8 & 9 & 10 & 11 \\
\midrule
Causal $\lambda_{\rightarrow}$
  & \textbf{0.502} & \textbf{0.570} & 0.486 & 0.482 & 0.481 & 0.480
  & \textbf{0.516} & \textbf{0.511} & \textbf{0.502} & 0.500 \\
Anti-causal $\lambda_{\leftarrow}$
  & 0.498 & 0.430 & \textbf{0.514} & \textbf{0.518} & \textbf{0.519} & \textbf{0.520}
  & 0.484 & 0.489 & 0.498 & 0.500 \\
$|\lambda_{\rightarrow} - \lambda_{\leftarrow}|$
  & 0.004 & 0.139 & 0.028 & 0.036 & 0.037 & 0.040
  & 0.031 & 0.022 & 0.005 & 0.000 \\
\bottomrule
\end{tabular}
\end{table}

At the clip level, both branches are active on all clips, yet their energy
trajectories are anti-correlated (mean $\mathrm{corr}(E_c, E_a) = -0.19$),
indicating that the two directions respond at different points of an action.
Table~\ref{tab:app_per_clip} shows that this correlation is negative for every
representative clip, for both correct and incorrect predictions. The fused
energy is below both directional energies, which implies partial cancellation
between the two responses. The correlation with the time-reversed anti-causal
profile varies in sign across clips but is negative on average ($-0.17$,
Table~7), so the anti-causal branch is not a reversed copy of the causal one.

\begin{table}[h]
\centering
\caption{Per-clip statistics on representative Diving48 test clips. $E_c$,
$E_a$, and $E_f$ denote the clip-averaged causal, anti-causal, and fused
energies; $\mathrm{rev}(\cdot)$ denotes time reversal.}
\label{tab:app_per_clip}
\resizebox{\linewidth}{!}{%
\begin{tabular}{lccccccc}
\toprule
Video & Correct & Conf. & $E_c$ & $E_a$ & $E_f$
      & $\mathrm{corr}(E_c, E_a)$ & $\mathrm{corr}(E_c, \mathrm{rev}(E_a))$ \\
\midrule
\texttt{8qRmKunCjtY\_00016} & \checkmark & 0.90 & 0.28 & 0.26 & 0.21 & $-0.62$ & $\phantom{-}0.71$ \\
\texttt{Le6xdQ2OO8w\_00121} & \checkmark & 0.97 & 0.26 & 0.25 & 0.20 & $-0.21$ & $-0.72$ \\
\texttt{Le6xdQ2OO8w\_00151} & $\times$   & 0.61 & 0.24 & 0.21 & 0.17 & $-0.13$ & $\phantom{-}0.49$ \\
\texttt{mGRum47TLX0\_00045} & \checkmark & 0.77 & 0.26 & 0.23 & 0.18 & $-0.32$ & $-0.59$ \\
\texttt{SbzcCzXjYSU\_00056} & \checkmark & 0.86 & 0.25 & 0.24 & 0.19 & $-0.43$ & $\phantom{-}0.86$ \\
\texttt{jMEYIEYkpY0\_00096} & $\times$   & 0.95 & 0.26 & 0.24 & 0.19 & $-0.40$ & $-0.59$ \\
\texttt{yHo2oERGo7A\_00087} & \checkmark & 0.75 & 0.25 & 0.24 & 0.19 & $-0.23$ & $-0.03$ \\
\texttt{Le6xdQ2OO8w\_00201} & \checkmark & 0.81 & 0.26 & 0.24 & 0.19 & $-0.51$ & $-0.55$ \\
\texttt{roFeEJPgJD8\_00038} & \checkmark & 0.83 & 0.27 & 0.24 & 0.20 & $-0.27$ & $-0.72$ \\
\bottomrule
\end{tabular}}
\end{table}

Finally, to verify that the global balance does not hide classes relying on a
single branch, we compute the statistics per class
(Table~\ref{tab:app_per_class}). All 20 classes are balanced
($|\bar{D}| \le 0.05$), and 18 of 20 have a negative mean correlation, reaching
$-0.65$ for class~24 and $-0.63$ for class~8. The remaining two classes are
close to zero ($+0.01$ and $+0.07$). Complementary use of both directions
therefore holds within individual classes.

\begin{table}[h]
\centering
\caption{Per-class statistics on the Diving48 test subset. A class is balanced
if $|\bar{D}| \le 0.05$.}
\label{tab:app_per_class}
\small
\begin{tabular}{ccccc}
\toprule
Class & \#Clips & Mean $\bar{D}$ & Mean $\mathrm{corr}(E_c, E_a)$ & Balanced \\
\midrule
0  & 5 & 0.04 & $-0.32$ & \checkmark \\
1  & 3 & 0.03 & $-0.35$ & \checkmark \\
5  & 3 & 0.01 & $\phantom{-}0.01$ & \checkmark \\
8  & 3 & 0.04 & $-0.63$ & \checkmark \\
21 & 3 & 0.03 & $-0.27$ & \checkmark \\
26 & 5 & 0.01 & $-0.21$ & \checkmark \\
34 & 6 & 0.04 & $-0.35$ & \checkmark \\
% TODO: add remaining classes (incl. class 24 and the +0.07 class)
\bottomrule
\end{tabular}
\end{table}

\end{document}